\documentclass[sigchi,nonacm]{acmart}
\usepackage{booktabs} 
\usepackage{subfig}
\usepackage{CJKutf8}
\newcommand{\chinese}[1]{\begin{CJK*}{UTF8}{gbsn}#1\end{CJK*}}

\begin{document}
\title{"Love is as Complex as Math": Metaphor Generation System for Social Chatbot}

\author{Danning Zheng}
\affiliation{
  \institution{University Of Rochester}
  \city{Rochester}
  \state{NY}
}
\email{dzheng2@u.rochester.edu}

\author{Ruihua Song}
\affiliation{
  \institution{Microsoft}
  \city{Beijing}
}
\email{song.ruihua@microsoft.com}

\author{Tianran Hu}
\affiliation{
  \institution{University Of Rochester}
  \city{Rochester}
  \state{NY}
}
\email{thu@cs.rochester.edu}

\author{Hao Fu}
\affiliation{
  \institution{Microsoft}
  \city{Beijing}
}
\email{fuha@microsoft.com}

\author{Jin Zhou}
\affiliation{
  \institution{Beijing Film Academy}
  \city{Beijing}
}
\email{whitezj@vip.sina.com}

\begin{abstract}
As the wide adoption of intelligent chatbot in human daily life, user demands for such systems evolve from basic task-solving conversations to more casual and friend-like communication. To meet the user needs and build emotional bond with users, it is essential for social chatbots to incorporate more human-like and advanced linguistic features. In this paper, we investigate the usage of a commonly used rhetorical device by human -- metaphor for social chatbot. Our work first designs a metaphor generation framework, which generates topic-aware and novel figurative sentences. By embedding the framework into a chatbot system, we then enables the chatbot to communicate with users using figurative language. Human annotators validate the novelty and properness of the generated metaphors. More importantly, we evaluate the effects of employing metaphors in human-chatbot conversations. Experiments indicate that our system effectively arouses user interests in communicating with our chatbot, resulting in significantly longer human-chatbot conversations.

\end{abstract}




\maketitle

\captionsetup[table]{position=below}

\section{Introduction}
\begin{figure}
\includegraphics[width=0.45\textwidth]{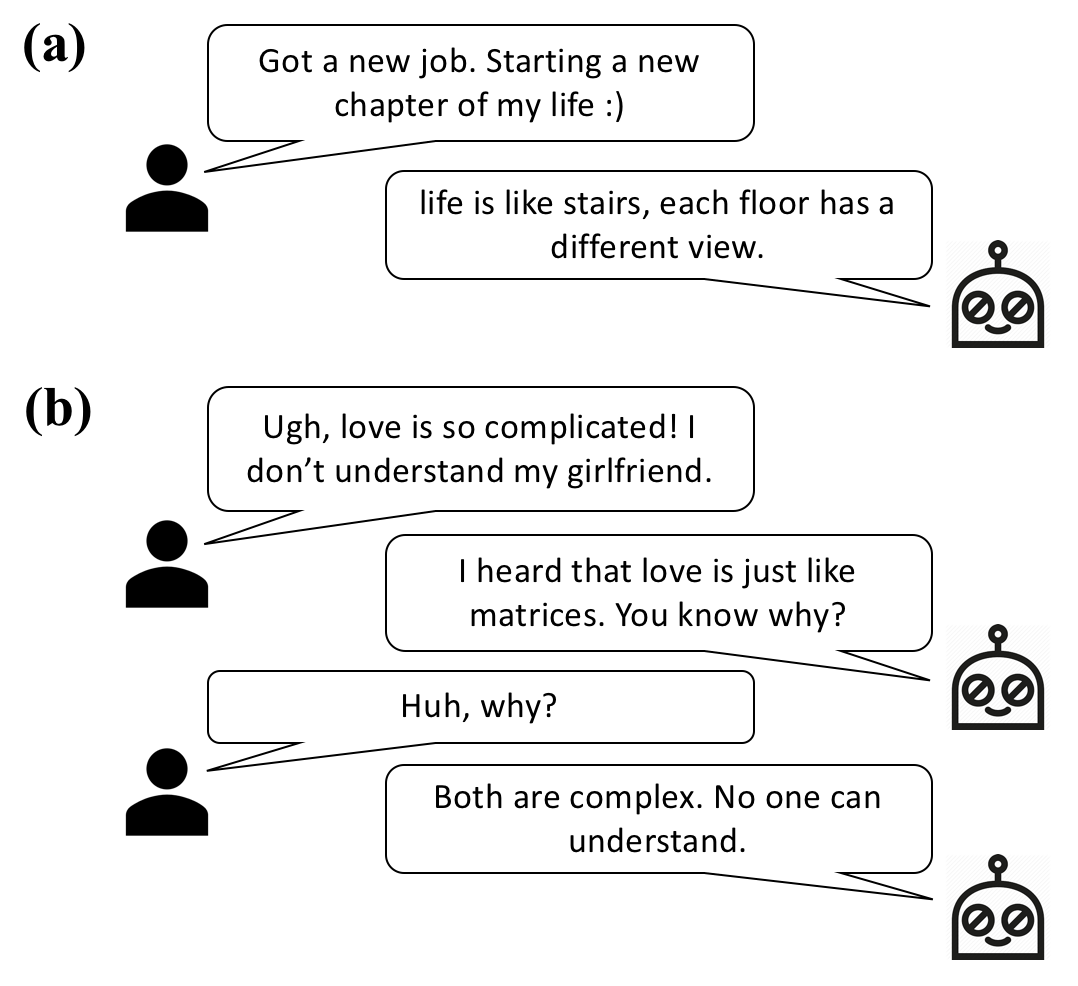}
\caption{Two examples of our social chatbot using metaphors in conversation with users. (a) demonstrates a one-round conversation in which the chatbot directly says the whole metaphor sentence. (b) demonstrates a two-round conversation in which the chatbot first says a novel comparison to interact with user, followed by the explanation in the second round.}
\label{fig:chat}
\end{figure}

In recent years, we have witnessed the emergence of a new type of automatic conversational system -- social chatbots, such as SimSimi \footnote{https://www.simsimi.com}, Microsoft XiaoIce \footnote{https://en.wikipedia.org/wiki/Xiaoice}, and Replika \footnote{https://replika.ai}. Different from traditional task-oriented bots~\cite{lopez2017alexa}, social chatbots are designed to "communicate" and build "emotional bonds" with users~\cite{eliza}. Social chatbots bring users closer and better engage them in human-computer conversations. As an illustrative example, Microsoft XiaoIce has been an extremely popular social chatbot since it was released in 2014. XiaoIce has accumulated 660 million users worldwide and on average users interact with XiaoIce 60 times a month\footnote{http://www.chinadaily.com.cn}. 

Meanwhile, the new purposes of social chatbots also introduce new challenges: speaking more like a "virtual friend" to users. Thus, social chatbots should be capable of handling more casual and open-domain conversations. Although much work has been done on chatbots~\cite{xu2017new,lopez2017alexa,li2016deep}, this work has mostly focused on task-oriented chatbots and on making chatbots talk "correctly" instead of "casually". To enrich the expressions of social chatbots, a natural approach is to introduce more human-like and advanced linguistic features. Figurative language is frequently used in human communication~\cite{lakoff}. Previous studies \cite{doi:sam,book:kaal} suggest that figurative language such as metaphors and sarcasm are key to interesting and engaging conversations. Furthermore, Roberts et al. \cite{doi:richard} examined the specific goals of people using figurative language in conversations and reported that most people view metaphors making conversations more interesting. Therefore, in this work, we develop a new social chatbot that conducts conversations with users using automatically generated metaphors. 

Our framework starts from a randomly selected target-source pair, such as "love" and "math". The system then quantitatively finds proper connections between source and target. For example, "(being) complex" is considered as a feature shared by both "love" and "math". Based on the target-source pair and the discovered connection, the framework generates metaphorical sentences (e.g. "Love is as complex as math.") We validated our system from two perspectives: 1) the quality of generated metaphors in terms of properness and creativity; and, more importantly, 2) how do users react to these metaphors in real human-computer conversations. For the first evaluation, human annotators are asked to label the quality of the generated metaphors. The results show that our framework is capable of generating novel and proper metaphors. Regarding the second evaluation, we study user experiences with the generated metaphors. We focus on metrics including but not limited to friendliness (i.e., how much the chatbot is speaking like a friend) and follow-up rate (i.e., the desire to respond to the chatbot). Test results indicate that users are more interested and are significantly more willing to respond when a chatbot uses metaphors. Metaphors also marginally significantly increased the perceived friendliness of chatbots. 

The contributions of this study are threefold:

1) We propose an automatic metaphor generation system for social chatbots. To the best of our knowledge, this is the first work that considers generating metaphors for conversational systems.

2) We conduct user studies to evaluate the quality of generated metaphors. Results show that the system is able to generate novel and proper metaphors. 

3) We systematically evaluate the effect of using metaphors in human-computer conversations. The results reveal that metaphors make users feel more interested and more willing to respond. 


\section{Related Work}
Metaphors (e.g., Love is like chocolates, sweet and bitter at the same time) are a figure of speech involving the comparison of one thing with another thing of a different kind and are used to make a description more emphatic or vivid \cite{richards1965}. 

Previous studies \cite{doi:sam,book:kaal} suggest that the use of figurative language such as metaphors and sarcasm are important for creating interesting and engaging conversations. Roberts et al. \cite{doi:richard} report that among all major figurative language types, metaphors are most able to make conversations more interesting. Specifically, 71\% of  participants indicated that they use metaphor to add interest to conversations and 12\% use metaphors to get attention from their conversational partner.

Early works \cite{xu2017new,lopez2017alexa,li2016deep} on human-computer conversation systems mainly focused on task-completion, such as customer service, making recommendations and answering questions. Researches on task-oriented systems are mainly focusing on addressing users' queries and generating informative answers. In recent years, more and more attention has been paid to non-task-oriented chatbots \cite{eliza}, which aim to hold casual and engaging conversations with users in open domains. A number of studies \cite{hu2018touch,huber2018emotional} have been done to meet users' emotional needs and make conversation systems more engaging. 

Despite the pervasiveness of figurative language in human conversations, little attention has been paid on integrating figurative language with chatbots. Inspired by Roberts et al.'s study \cite{doi:richard}, we propose a metaphor generation system that is capable of generating metaphors and enhancing users' engagement with chatbot systems.

\begin{table}[htbp]
\begin{tabular}{|l||c|} \hline\hline
\textbf{Target} & \textbf{Freq.} \\ \hline
Parting & 0.68\% \\
Love & 0.38\% \\
Heart & 0.21\% \\
World & 0.20\% \\ 
Mother & 0.16\% \\
Beauty & 0.12\% \\
Man & 0.11\% \\
Dream & 0.10\% \\
Life & 0.10\% \\
Happiness & 0.09\% \\
\hline\hline
\end{tabular}
\quad
\begin{tabular}{|l||c|c|} \hline\hline
\textbf{Source} & \textbf{Freq.} & \textbf{Conc. R.} \\ \hline
Food & 0.92\% & 4.80 \\
Signal & 0.28\% & 3.86 \\
Game & 0.27\% & 4.50 \\
Father & 0.22\% & 4.52 \\ 
Robot & 0.21\% & 4.65 \\
Wife & 0.20\% & 4.13 \\
Picture & 0.17\% & 4.52 \\
Brother & 0.16\% & 4.43 \\
Photos & 0.16\% & 4.93 \\
Phone & 0.15\% & 4.86 \\
\hline\hline
\end{tabular}
\caption{Top 10 most frequent abstract concepts and concrete concepts in our chatbot conversation log. \textit{Conc. R.} is the abbreviation for concreteness rating.}
\label{tbl:frequency}
\end{table}

\section{Metaphor Generation System}

\begin{figure*}[t]
\centering
\subfloat[Adjective]{
\includegraphics[width=.33\textwidth]{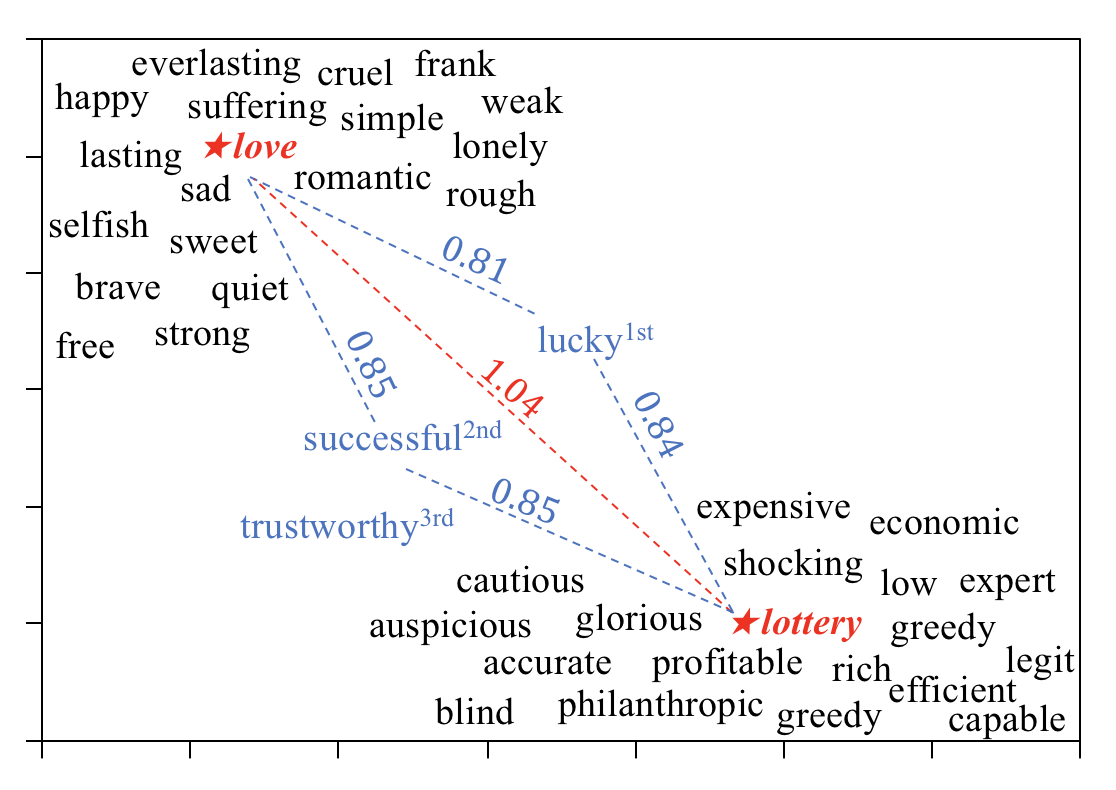}
}
\subfloat[Verb]{
\includegraphics[width=.33\textwidth]{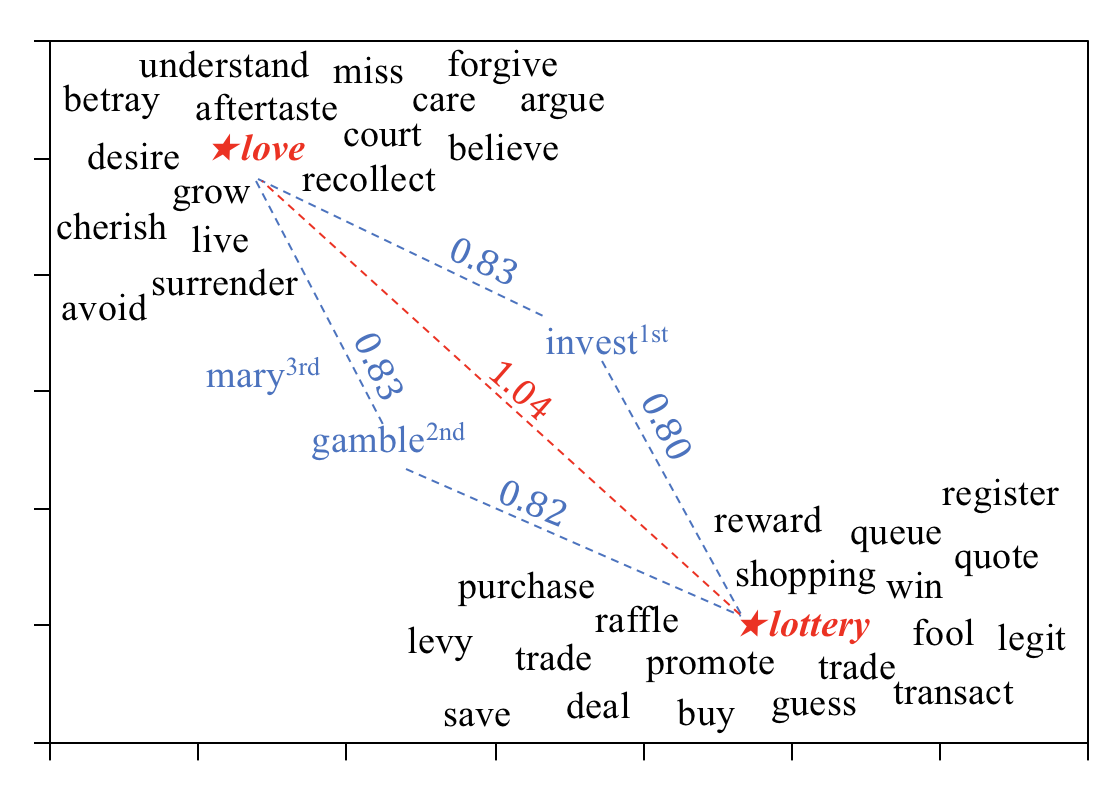}
}
\subfloat[Noun]{
\includegraphics[width=.33\textwidth]{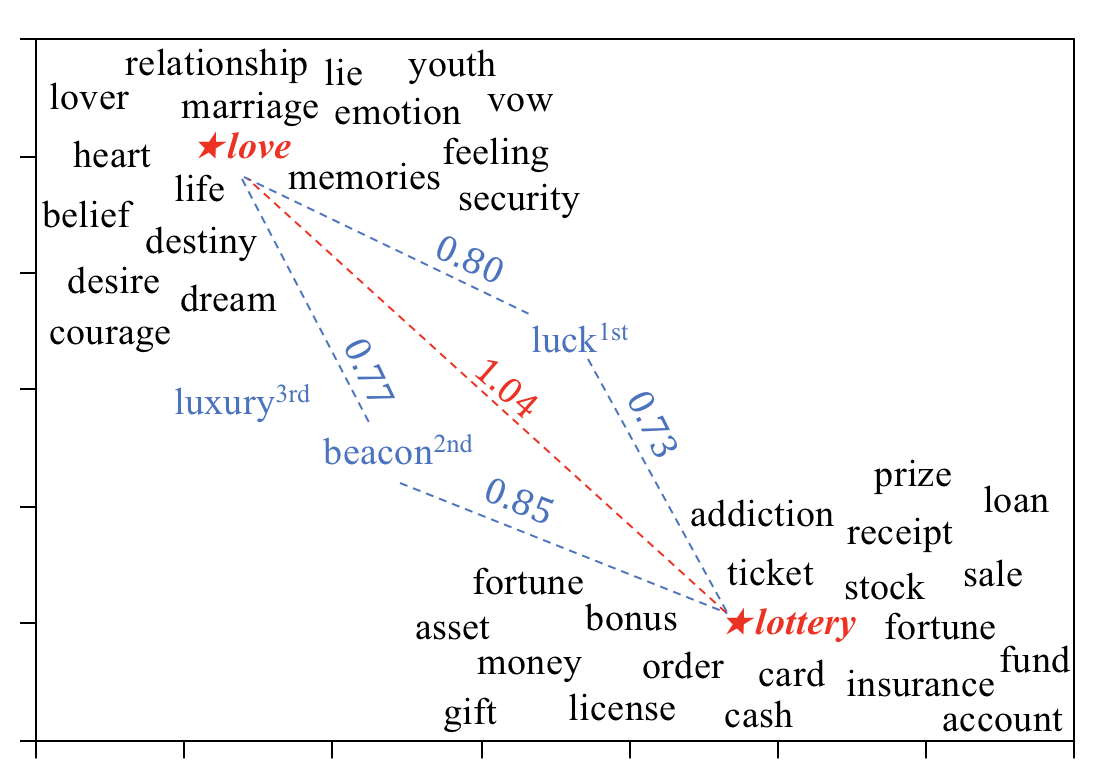}
}
\caption{An illustration of the connecting words (in \textcolor{blue}{blue}) for target \textit{love} and source \textit{lottery} (in \textcolor{red}{red}) by different part of speech (POS) tags. Plots (a), (b), and (c) respectively show adjectives, verbs, and nouns in the underlying word vector space. Numbers on the dotted lines represent the semantic distance (defined as $1 - cosine(\cdot,\cdot)$) between a pair of words. }
\label{fig:connect_word}
\end{figure*}

\subsection{Targets and Sources Selection}
Previous cognitive linguistic studies show that target and source are usually of different types of concepts: target are usually from abstract domains, while sources from concrete domains~\cite{lakoff}. In other words, by utilizing metaphors, people manage to explain and express less-understood and abstract concepts (i.e., targets) using well-understood and concrete concepts (i.e., sources)~\cite{analogy}. Therefore, to select suitable targets and sources, we applied two different approaches in our system.

To select targets, we first followed previous linguistic studies and collected 122 poetic themes~\cite{Gagliano,Gero2018challenges}. Please note that poetic themes are usually abstract concepts, which makes them ideal candidates for targets. We then extended the candidate set by adding the closest five concepts of each poetic theme\footnote{The closest concepts are extracted from a pre-trained word embedding space learned from millions of social media (Weibo) posts}. To ensure that the concepts are actually being used in human-computer conversations, we further analyzed the frequency of each concept in our chatbot conversation log. We filtered out the concepts that are rarely used (frequency lower than 0.001\%) and obtained 96 concepts. These concepts were used as target candidates in our system. These concepts spanned many diverse topics, such as romance (e.g., "love", "heart"), history (e.g., "war", "peace"), and nature (e.g., "earth", "spring"). Table~\ref{tbl:frequency} shows the top ten most popular concepts, as well as their frequencies.

To select sources, we considered two factors of a concept: popularity in human-computer conversations and concreteness. We first extracted the top 10,000 frequently used nouns from our chatbot conversation log. We then learned the concreteness scores for these words from a concreteness database introduced by Brysbaert et al~\cite{brysbaert2014concreteness}. The database assigns concreteness ratings for 40 thousand English words, and the ratings evaluate the degree to which the concept denoted by a word refers to a perceptible entity~\cite{brysbaert2014concreteness}. We took the most concrete 3,000 nouns as source candidates for our system. Table~\ref{tbl:frequency} also shows the top ten most concrete concepts and their scores. 

\subsection{Discover Connections between Targets and Sources}


Besides a target and a source, a metaphor also requires a connection between these two concepts. The connection is usually an expression, which is not only semantically close to both the target and source, but also maintains a balanced semantic distance to the two words~\cite{Gagliano}. In our framework, we quantitatively discovered words linking targets and sources semantically, and refer to these words as \textit{connecting words}.


We first located targets and sources in a word embedding space. Since the distance in the space represents the semantic similarities of words~\cite{mikolov2013linguistic}, we can quantify how good a word is in terms of connecting target and source from two perspectives: 1) connectivity: the semantic distances from a connecting word to target and source should be smaller than the semantic distance between target and source; and 2) balance: a connecting word should maintain a balanced distance to target and source, thus drawing the target and source together. These two aspects can be clearly visualized in Figure~\ref{fig:connect_word}. For example, in Figure~\ref{fig:connect_word} (a), \textit{lucky} is a connecting word that demonstrates both connectivity and balance between target \textit{love} and source \textit{lottery}. Therefore, combining these two aspects, we designed a \textit{connecting score}. Formally, given a target $T$ and a source $V$, the connecting score of a word $X$ for $T$ and $V$ is defined as:


\begin{equation}
\begin{split}
connecting(X|T,V) = dist(T,X) + dist(V,X) + \\ 
log(|dist(T,X) - dist(V,X)| + \beta)
\end{split}
\end{equation}

The lower the connecting score, the better a word could link the two concepts. For all the target-source pairs, we ranked all words according to their \textit{connecting score} in ascending order and choose the top 5 words as the connecting words. 

\subsection{Identify Similarities by Different POS }
As connecting words should convey enough information, we considered content words (i.e., adjectives, verbs, and nouns) as candidates. Connecting words semantically links a target and a source, but connecting words of different part of speech (POS) links the two concepts in different ways. We summarize the most representative case in each category: 1) The connecting word is an adjective and it is a common attribute or property shared by the target and the source. For example, "complex" is a proper connecting word for target "love" and source "math". 2) The connecting word is a verb and it can modify both the target and the source. For example, "scream" is a proper connecting word for target "soul" and source "football fans". 3) The connecting word is a noun and its relationship with the target is the same as its relationship with the source. For example, "gamble" is a proper connecting word for target "love" and source "lottery".

We designed different methods for generating metaphor from connecting words of different POS. Table \ref{tbl:example_sentences} reports example metaphors generated from (target, source, connecting word) triplets, and also shows the POS tags of connecting words. In the next three sections, we report our approach for each of the three categories. 

\begin{figure}
\includegraphics[width=0.4\textwidth]{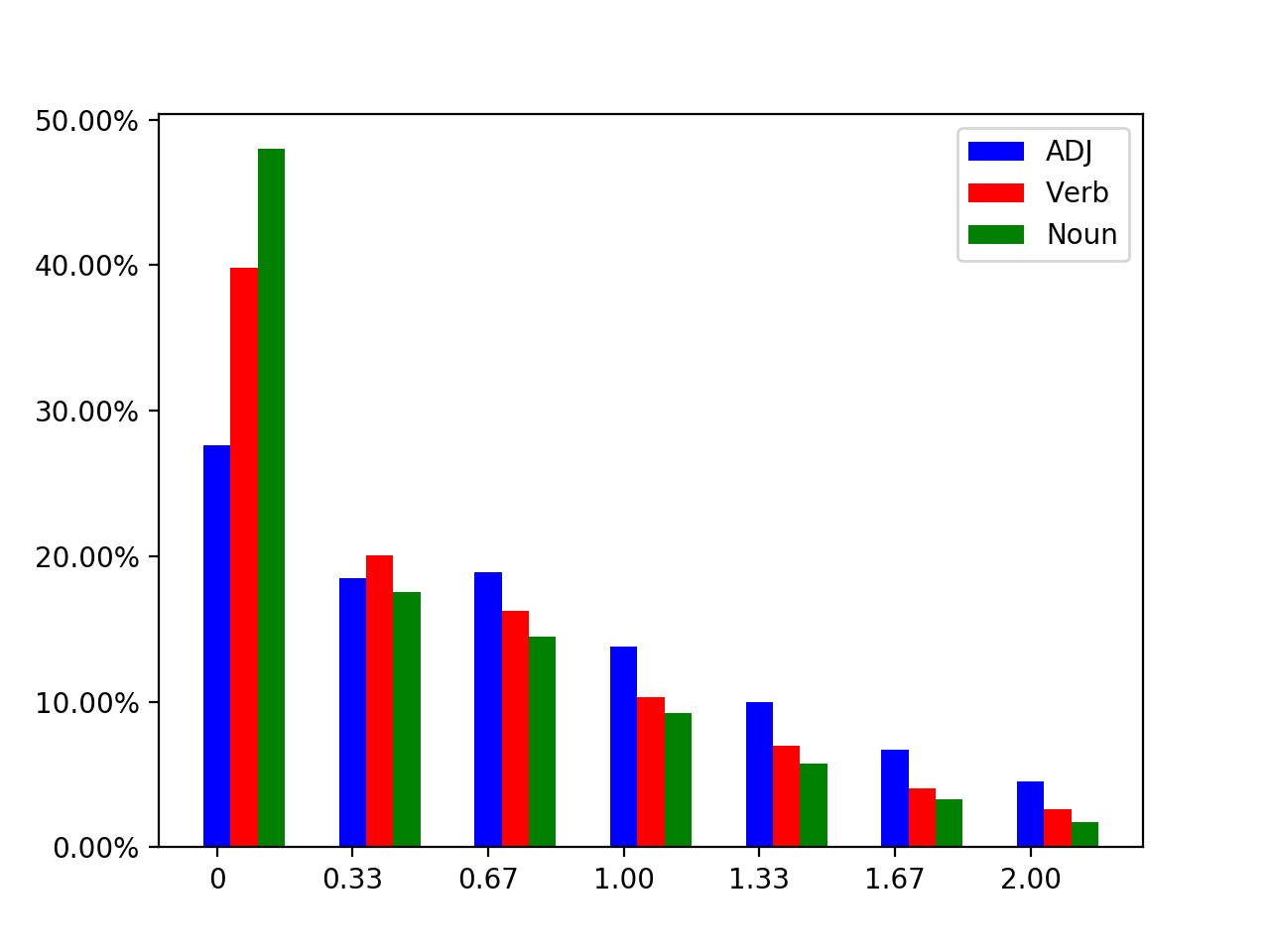}
\caption{The properness score distribution of connecting words by different POS categories. X-axis shows the properness scores, ranging from 0 (not proper) to 2.0 (very proper). Y-axis shows the percentage. }
\label{fig:connectword_result}
\end{figure}

\begin{figure*}[t]
\centering
\subfloat[smoothness]{
\includegraphics[width=.33\textwidth]{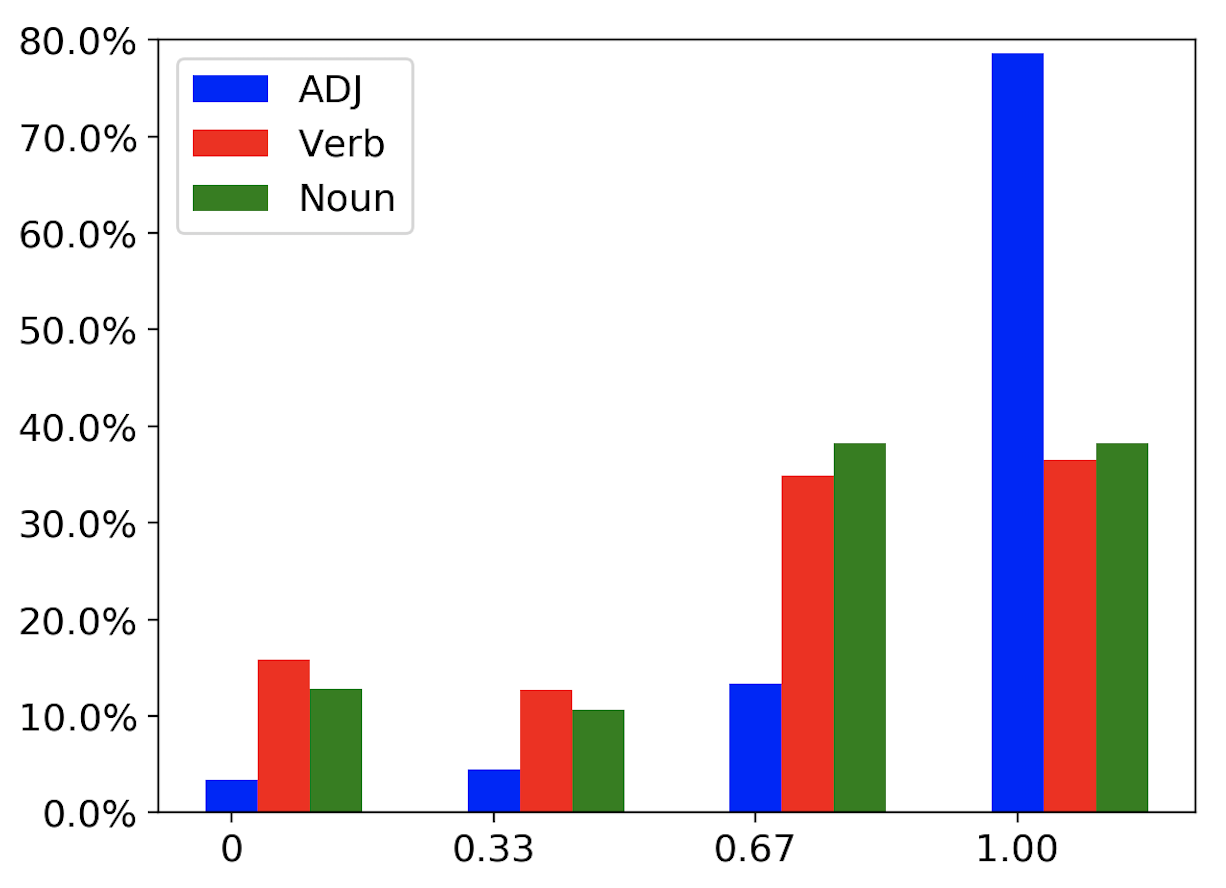}
}
\subfloat[properness]{
\includegraphics[width=.33\textwidth]{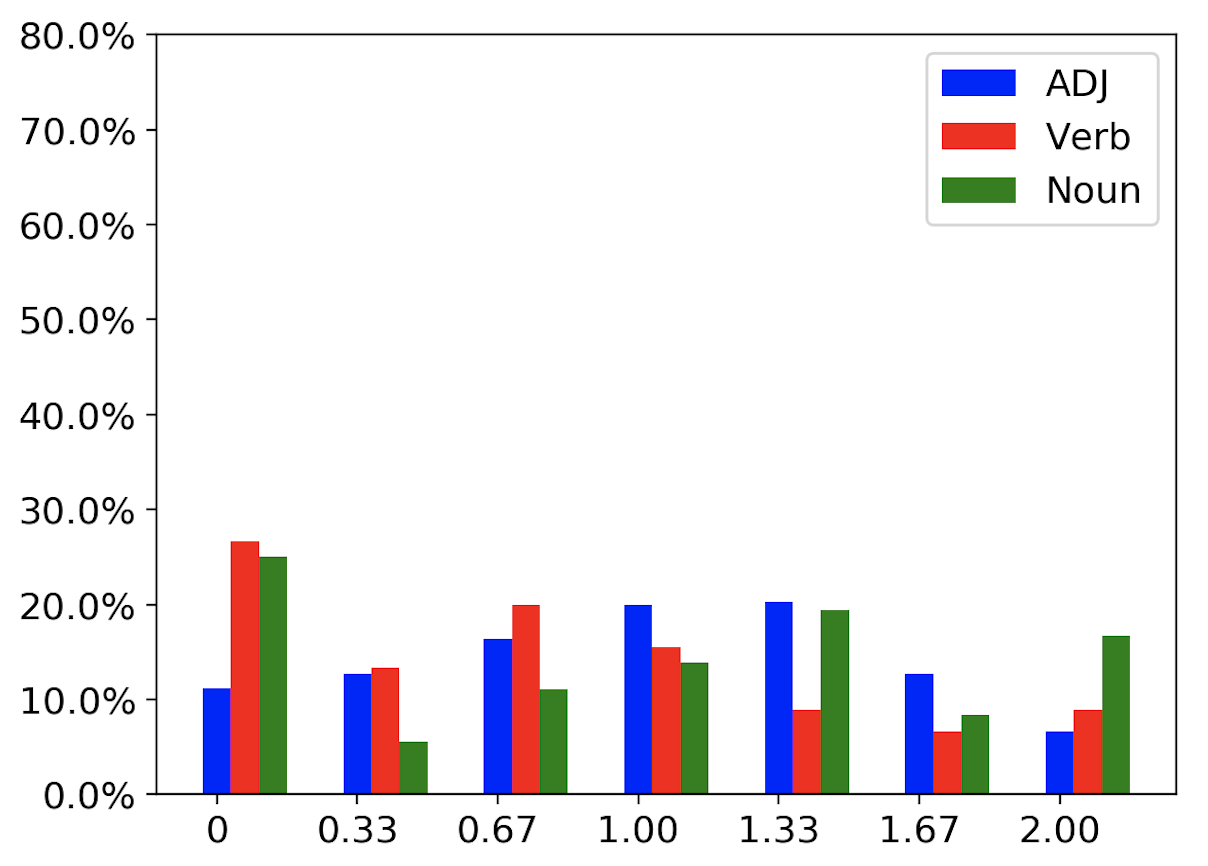}
}
\subfloat[novelty]{
\includegraphics[width=.33\textwidth]{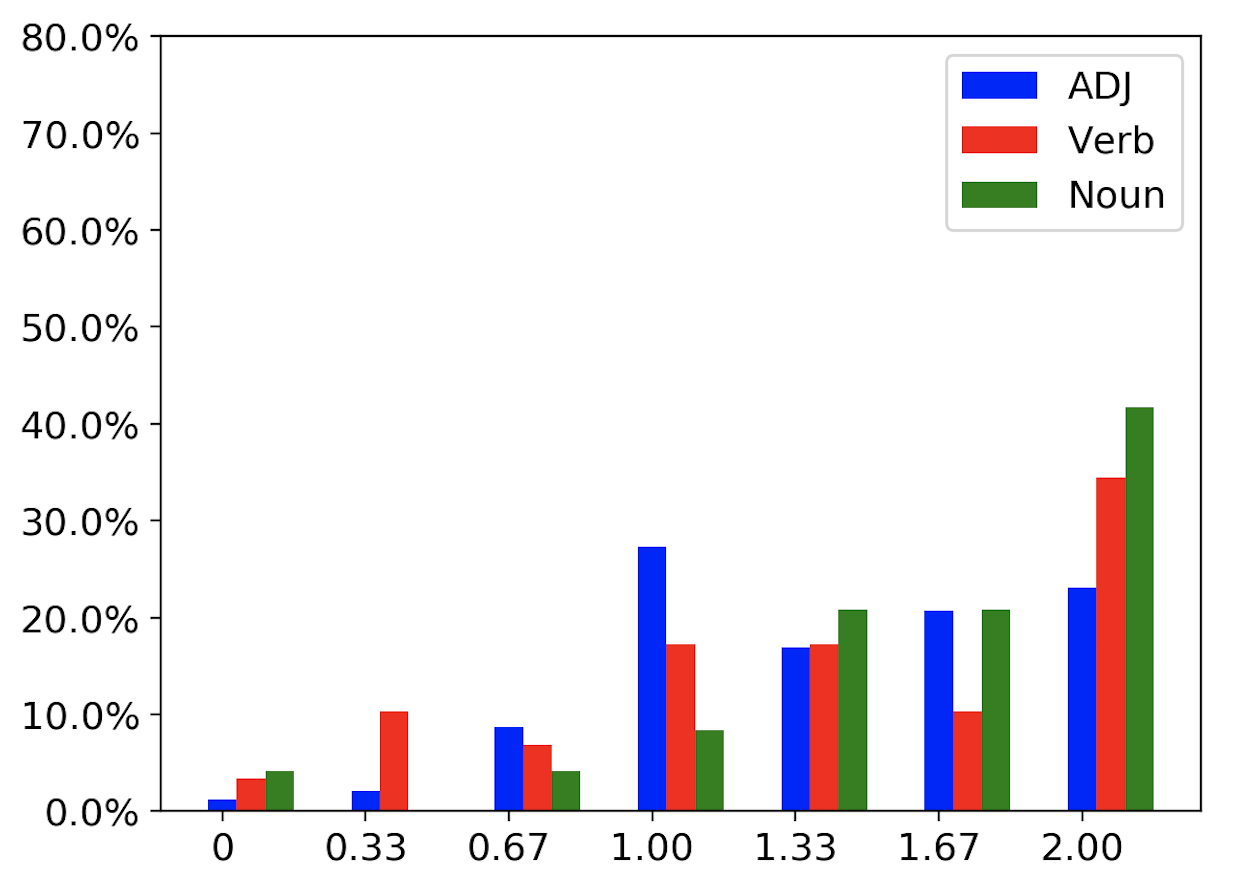}
}
\caption{Plots (a), (b), and (c) respectively show the smoothness score, properness score, and novelty score distribution of metaphor sentences by different categories of connecting words: adjective, verb, and noun.}
\label{fig:metaphor_result}
\end{figure*}

\subsection{Generate Metaphors with ADJ Connecting Words}
Ortony et al. \cite{ortony1979} argue that metaphors project high-salience properties of a descriptive source term (the source) onto a target term (the target) for which those properties are not already salient. In other words, a proper connecting word for a target-source pair can be 1) used to describe the target, and 2) a salient attribute of the source. Note that condition 2 is more restrictive than condition 1: both "sour" and "sweet" can be used to describe apples, but only "sweet" is a salient attribute. We validated adjective connecting words based on these two conditions by checking if people have used the adjective to describe the target and the source before.

Specifically, to validate condition 1, we send two queries \textit{adjective T} (e.g. "sweet love") and \textit{T is adjective} (e.g. "love is sweet") to a web search engine and recorded the total number of returned web pages. Similarly, to validate condition 2, we queried \textit{as adjective as (a|an) V} (e.g. "as sweet as apples"), and recorded the number of returned web pages. We considered an adjective as proper if both conditions are satisfied, i.e., both numbers are larger than certain thresholds. To generate complete metaphor sentences, we then manually constructed a few templates: \textit{T is adjective, just like V.}, \textit{T is as adjective as (a|an) V.}, and \textit{T is like (a|an) adjective V.} 

\subsection{Generate Metaphors with Verb Connecting Words}
A key observation is that verb and noun \textit{connecting words} tend to exhibit diverse relationships with targets and sources. Therefore, we do not identify and handle all possible relationships, but rather handle the most representative case: subject and verb associations. Subject-verb is the most fundamental sentence structure. It is also an effective feature in metaphor detection \cite{shutova2010metaphor} and metaphor generation \cite{figure8}. Thus, to validate whether a verb exhibits the same relation with a target and source pair, we verified if target-verb and source-verb each demonstrate subject-verb relations.  

Starting from a verb connecting word and a pair of target T and source V, we sent two queries \textit{verb + T} and \textit{verb + V} to a search engine and retrieved the top 10,000 web snippets for each query. After removing duplicates, we had 615.3 web snippets on average for each keyword pair. We then filtered invalid sentences (e.g., broken sentences, advertisements, and sentences that don't contain both keywords), which resulted in 4.6 sentences on average for each keyword pair. We analyzed the syntactic dependency structure of each sentence and filtered out those sentences in which the target-verb or source-verb relation was not subject-verb. We ranked all sentences of targets by their semantic distance to the source word, in which the distance was calculated as the average distance of every word (excluding stopwords) in the sentence to the source word. We used the sentence with smallest distance as the explanation and generated \textit{T is like V, [explanation].} metaphors. 

\subsection{Generate Metaphors with Noun Connecting Words}
Similar to the verb case, noun connecting words exhibit diverse relations with targets and sources. Therefore, we  identified and handled the most representative relation: subject-predicate-object patterns. The idea is that if there exists a certain predicate such that both target-predicate-noun and source-predicate-noun frequently occur as subject-predicate-object in a large text corpus, then we know \textit{predicate + noun} is a phrase that can modify both target and source. 

We followed the same procedure to collect sentences for targets and sources from the search engine and filtered the sentences. On average, we collected 612.4 web snippets and 5.3 valid sentences for each keyword pair. We identified subject-predicate-object structure in each sentence via dependency parsing. We then followed the same approach to generate \textit{T is like V, [explanation].} metaphors.


\begin{table*}[htbp]
\begin{tabular}{|l||c|c|c|c|}
\hline
\textbf{Generated Metaphor} & \textbf{Smooth} & \textbf{Proper} & \textbf{Novel} & \textbf{POS} \\ \hline
\textcolor{red}{Time} is \underline{sweet}, like a \textcolor{orange}{tangerine}. & 1.0 & 2.0 & 1.7 & Adj. \\
\chinese{\textcolor{red}{时光}\underline{很甜}, 就像\textcolor{orange}{柑橘}一样。} & & & & \\
\hline

\textcolor{red}{Love} is like \textcolor{orange}{salary}, has a goal and is not \underline{blind}. & 1.0 & 2.0 & 1.7 & noun \\
\chinese{\textcolor{red}{爱情}就像\textcolor{orange}{工资}，都是有点目标的，不能\underline{盲目}。} & & & & \\
\hline

\textcolor{red}{Relationship} is like a \textcolor{orange}{park}, needs to be operated and \underline{maintained}.  & 1.0 & 2.0 & 1.0 & verb \\
\chinese{\textcolor{red}{感情}就像\textcolor{orange}{园区}，需要经营和\underline{维护}。} & & & & \\
\hline

\textcolor{red}{Loneliness} is like an \underline{empty} \textcolor{orange}{station}. & 1.0 & 1.3 & 0.7 & Adj. \\
\chinese{\textcolor{red}{孤独}就像\underline{空无一人}的\textcolor{orange}{车站}。} & & & & \\
\hline

\textcolor{red}{Soul} is like a \textcolor{orange}{football fan}, silently \underline{screaming}. & 0.7 & 1.0 & 1.0 & verb \\
\chinese{\textcolor{red}{灵魂}就像\textcolor{orange}{球迷}一样，在无声地\underline{呐喊}。} & & & & \\
\hline

\textcolor{red}{Marriage} is like a \textcolor{orange}{guide}, has its own set of \underline{rules}. & 0.7 & 1.0 & 0.0 & noun \\
\chinese{\textcolor{red}{婚姻}就像\textcolor{orange}{指南}，有自己的一套\underline{法则}。} & & & & \\
\hline

\textcolor{red}{Childhood} is very \underline{cute}, like a \textcolor{orange}{dolphin}. & 0.7 & 0.7 & 1.0 & Adj. \\
\chinese{\textcolor{red}{童年}很\underline{可爱}，就像\textcolor{orange}{海豚}一样。} & & & & \\
\hline

\textcolor{red}{Work} is as \underline{outstanding} as \textcolor{orange}{ballet}. & 0.7 & 0.0 & 0.0 & Adj. \\
\chinese{\textcolor{red}{工作}像\textcolor{orange}{芭蕾}一样\underline{出色}。} & & & & \\
\hline

\textcolor{red}{Life} is like a \textcolor{orange}{stairway}, has a \underline{direction}, not confused. & 0.3 & 0.0 & 0.0 & noun \\
\chinese{\textcolor{red}{人生}就像\textcolor{orange}{楼梯}，有\underline{方向}，再难不迷茫。} & & & & \\
\hline

\textcolor{red}{Time} is like \textcolor{orange}{sportswear}, will not \underline{fade}, memory will shine. & 0.0 & 0.0 & 0.0 & verb \\
\chinese{\textcolor{red}{时光}就像\textcolor{orange}{运动服}，不会\underline{褪色}，记忆也会发光。} & & & & \\
\hline

\hline
\end{tabular}
\caption{Examples of generated metaphors in decreasing order of the smoothness, properness, and novelty scores. Targets (in \textcolor{red}{red}), sources (in \textcolor{orange}{orange}), and connecting words (underlined) are highlighted.}
\label{tbl:example_sentences}
\end{table*}

\section{Metaphor Generation System Evaluation}
\subsection{Connecting Words Evaluation}
From all 96$\times$3000 pairs of target and source, we randomly sampled 500 pairs and used equation (1) to retrieve the top 5 adjective connecting words, verb connecting words, and noun connecting words, respectively. Each of the 500$\times$15 samples was annotated on a 3-point scale: 0 (not proper), 1 (proper), and 2 (very proper). Each sample was labeled by 3 human judges, and its average was used as the final rating.  

Figure \ref{fig:connectword_result} shows the properness score distribution of connecting words by different POS categories. If we consider a connecting word with score $>=1$ as proper, then overall we have 1965 (26.2\%) proper connecting words, consisting of 847 adjectives, 597 verbs, and 521 nouns. An important observation is that adjective connecting words achieve higher scores than verb connecting words and verb connecting words achieve higher scores than noun connecting words. This result aligns with our previous analysis that verb and noun connecting words exhibit more diverse relations with targets and sources. In the next section, we evaluate the metaphors generated using the 1965 proper connecting words.

\subsection{Generated Metaphors Evaluation}
Metaphor generation was evaluated with 1965 proper (target, source, connecting word) triplets. We were able to generate a total of 461 metaphor sentences: 351 with adjectives, 63 with verbs, and 47 with nouns. The main reason for why there were fewer metaphors with verbs and nouns than with adjectives is that we only handled the subject-verb and subject-predicate-object patterns, though other possible relations exists. 

Each generated sentence was evaluated by three human annotators from the following perspectives: 1) smoothness: if a sentence is clear and grammatically correct; 2) properness: if the comparison and explanation are understandable and make sense; and 3) novelty: if the comparison or explanation is fresh, novel or surprising. Smoothness is a binary metric: 0 (not smooth) or 1 (smooth); properness and novelty are annotated on a 3-point scale: 0 (not at all), 1 (moderately), and 2 (very strongly). These three metrics reflect progressive relationships, i.e., a metaphor cannot be proper unless the sentence is clear, and the novelty score is only meaningful if the comparison is proper. We use the average of three annotators as the final score. Table \ref{tbl:example_sentences} reports the assigned scores of these three metrics for the example metaphors. A sentence with a smoothness score $\geq$0.7 is considered clear, and a metaphor with a properness score $\geq$1 is considered proper. We generated 330, 45, 36 smooth sentences and 242, 29, 24 proper sentences with adjectives, verbs, and nouns, respectively.

Figure \ref{fig:metaphor_result} shows the score distribution of each metric by different categories of connecting words. From plot (a) we can see that 92.02\% of metaphors generated with adjectives are smooth while only 71.43\% and 76.6\% of the metaphors generated with verbs and nouns, respectively, are smooth. This result is not surprising since the former approach applies a far more restrictive template than the latter two approaches. Plot (b) shows similar distributions for different word categories. However, it is still worth noting that 20\% more metaphors generated with verbs are inappropriate as compared to the metaphors generated with adjectives and nouns. This is probably because subject+verb is a relatively loose constraint. Finally, plot (c) reveals interesting differences in the distributions: the novelty distribution of adjective metaphors is evenly distributed between 1.00 and 2.00, while the novelty distribution of noun metaphors is continuously increasing from 1.00 to 2.00. There are 11.3\% and 18.5\% more strongly novel metaphors (i.e., novelty score $\geq$2.0) with verbs and nouns than with adjectives. This is because noun and verb sentences are longer and tend to provide richer explanations.  

\section{User Study}
\subsection{Evaluation Metrics}
Although several metrics have been proposed in previous works to evaluate the performance of chatbots \cite{li2016deep,xu2017new}, these metrics are mostly designed for task-oriented conversations. The main focuses of social chatbots, such as user experience in casual conversations, are usually ignored in these existing metrics. Therefore, we propose that one reasonable way to evaluate social chatbots would be to align the metrics with the goals of social chatbots: to build emotional bonds with users and become their friend. Therefore, we designed the following metrics to evaluate our system: 
(1) dialogue quality: do users think that the chatbot generates meaningful and informative content?; (2) friendliness: does the content generated by the chatbot make users feel that the chatbot is personable or engaging in the way a friend would?; (3) follow-up rate: does the content generated by the chatbot make users want to respond and keep the conversation going?

\subsection{Experiment Design}
Our data consists of 52 randomly sampled metaphor sentences with properness score $\geq$1. We implemented two different approaches to integrate our metaphor generation system into a chatbot. In the first approach, a chatbot directly says the complete metaphor sentence, e.g., "Heart is shining like a diamond". In the second approach, a chatbot first says a comparison, e.g. "I heard that heart is like a diamond. Do you know why?", and then follows with the explanation in the second round, e.g. "Because both are shining." Both approaches were compared to the baseline where a chatbot simply says the literal sentence, e.g., "Heart is shining." Therefore, for each metaphor sentence, we generated three different types of expressions: one-round metaphor, two-round metaphor, and literal sentence. 

We recruited three annotators and assigned them to rate all 50 sentences for all three types of expressions. Each metric was rated on a 5-point scale, from strongly disagree (-2) to strongly agree (2). To deal with the dependency of our within-subject experiment design, we first used a repeated measures ANOVA to analyze the differences among group means. We then performed Tukey post-hoc tests to compare all the group means in pairs and reported the significances of the pairwise differences.

\subsection{Statistical Result}
The results of our statistical tests are reported in Table \ref{tbl:user_study}. There is a statistically significant effect of expressions on dialogue quality and follow-up rate as determined by one-way repeated measures ANOVA. There is also a marginally significant difference in the friendliness score. The results suggest that integrating metaphors with a conversation system helps to attract users and makes conversations more interesting. This result aligns with prior studies on human-human conversations\cite{book:kaal,doi:richard}.

A Tukey post hoc test revealed that dialog quality (\textit{p} value = 0.013) and follow-up rate (\textit{p} value = 0.001) were  significantly higher when a chatbot directly stated the metaphor as compared to a literal sentence. There was also a marginally significant increase in friendliness score if a chatbot said a metaphor instead of a literal sentence (\textit{p} value = 0.09). However, there were no statistically significant differences between saying a metaphor in two-round conversations and saying a literal expression for dialogue quality, friendliness, and follow-up rate (\textit{p} value = 0.36, 0.42, and 0.18, respectively). One possible explanation is that user study subjects might feel that two-round conversations are unnecessary.

\begin{table}[htbp]
\begin{tabular}{llll} 
\toprule
 & \textbf{Dialog. Q.} & \textbf{Friendliness} & \textbf{Follow R.} \\ 
\midrule
1-round metaphor & $3.60\pm0.75$ & $3.55\pm0.67$ & $3.58\pm0.70$ \\
2-round metaphor & $3.45\pm0.88$ & $3.41\pm0.80$ & $3.40\pm0.84$\\
literal sentence & $3.32\pm0.84$ & $3.34\pm0.75$ & $3.22\pm0.80$ \\
\midrule
\textbf{\textit{F}-statistic} \\ 
\multicolumn{1}{r}{\textit{F}-value} & 3.719** & 2.417*  & 6.280** \\
\multicolumn{1}{r}{\textit{p}-value} & .031 & .09 & 0.004 \\
\midrule
**\textit{p}<.05, *\textit{p}<.1
\end{tabular}
\caption{Means and standard deviations of dialogue quality, friendliness, and follow-up rate scores for each type of expression in our user study. The F-statistic from a repeated measures ANOVA is also shown for each metric.}
\label{tbl:user_study}
\end{table}

\section{Testing on Real Chatbot Users}
To more robustly evaluate the effect of our system, we integrated metaphors with an existing social chatbot and analyzed how different expressions affect real chatbot users' follow-up rate. As users were unaware of the ongoing test, we were able to eliminate any potential biases. 


When integrating metaphors with social chatbot systems, we sought to make the integration context-aware and fit in the conversation flow, i.e., the metaphor is relevant to the conversation topic and the metaphor matches users' input. For example, if a user is talking about their boyfriend or girlfriend, a metaphor for love or marriage could be a good fit in the conversation. We used question-answer relevance, keyword matching, and topic similarity as input features and trained a classifier to predict whether a metaphor should be triggered. \footnote{Details of this classifier are beyond the scope of this paper and were omitted.} When a metaphor was a good fit in the conversation flow, we randomly triggered one of the three expressions. 

We tested our system on 924 users within a 3 week period. Users' follow-up rates are 22\%, 27\%, and 41\% for literal sentences, one-round metaphors, and two-round metaphors, respectively. Overall, the results show that both metaphor expressions achieve more follow-ups than literal expressions. Importantly, we found that the follow-up for two-round metaphors was the highest among all three expressions, which is contradictory to the findings from our user study. One possible explanation is that in the user study, two-round conversation might seem weird to annotators because there is no conversation context, and thus annotators assign lower scores. However, in real human-computer conversations, users prefer more interaction with a chatbot. Sample dialogues are presented in Table \ref{tbl:user_study_sample}.

\begin{table}
\begin{tabular}{rl}
\hline
\hline
Speaker & Utterance \\
\hline
Chatbot & Tell me what is your favorite food? \\ 
& (\chinese{说说你喜欢吃的东西是什么啊？})\\
User & Beef (\chinese{牛肉})\\
Chatbot & Only beef, without noodles?\\
& (\chinese{只要牛肉不要面？})\\
& You know what, \textcolor{red}{life} is like \textcolor{orange}{chicken}.\\
& (\chinese{你知道吗，\textcolor{red}{生活}就好比是\textcolor{orange}{鸡肉}})\\
User & Why? (\chinese{为什么？})\\
Chatbot & \underline{Mild}. (\chinese{\underline{清淡}})\\
User & Chicken may not be mild. (\chinese{鸡肉不清谈})\\
\hline
Chatbot & I am a cute girl. Believe it or not. \\
& (\chinese{我是萌妹纸，爱信不信}) \\
User & You are a girl. (\chinese{你是女孩子}) \\
Chatbot & I am so cute and definitely not a boy. \\
& (\chinese{这么可爱一定不是男孩子~}) \\
& A \textcolor{red}{child} is like a \textcolor{orange}{bowl}. Do you know why?\\
& (\chinese{\textcolor{red}{孩子}就像\textcolor{orange}{碗}，你猜为什么}) \\
User & Because your are fragile. \\
& (\chinese{因为你一碰就碎}) \\
Chatbot & \underline{Innocent}. (\chinese{\underline{干净}}) \\
User & You are like an unwashed bowl. \\
& (\chinese{你就像一个几天没洗的碗}) \\
\hline
\hline

\end{tabular}
\caption{Sample dialogues with two-round metaphors.}
\label{tbl:user_study_sample}
\end{table}

\section{Discussion and future work}
In this paper, we propose computational approaches to generate metaphors and report the effect of our system in the context of human-computer conversations. According to human evaluation results, our system is able to generate proper and novel metaphors. User study evaluations show that people feel that metaphorical expressions are more meaningful and interesting as compared to literal expressions. More importantly, integrating metaphors with an existing social chatbot increased users' follow-up rates. 

There are many interesting and valuable directions for future work, including studying how the properness and novelty of metaphors affects users' experiences and engagement in human-computer interactions. In the meantime, it will also be important to study possible improvements to our proposed metaphor generation model, such as enhancing the percentage of proper metaphors.


\bibliographystyle{ACM-Reference-Format}
\bibliography{sample-sigchi}

\end{document}